# Socially-Aware Shared Control Navigation for Assistive Mobile Robots in the Built Environment

Yifan Xu[1], Qianwei Wang[2], Vineet Kamat[3], and Carol Menassa[4]

[1]Ph.D. Student, Department of Civil and Environmental Engineering, University of Michigan, Ann Arbor, MI, USA Email: yfx@umich.edu

[2]Undergraduate Student, College of Literature, Science, and the Arts, University of Michigan, Ann Arbor, MI, USA Email: qweiw@umich.edu

[3]Professor, Department of Civil and Environmental Engineering, University of Michigan, Ann Arbor, MI, USA Email: vkamat@umich.edu

[4]Professor, Department of Civil and Environmental Engineering, University of Michigan, Ann Arbor, MI, USA Email: menassa@umich.edu

**ABSTRACT**

As the number of Persons with Disabilities (PWD), particularly those with one or more physical impairments, increases, there is an increasing demand for assistive robotic technologies that can support independent mobility in the built environment and reduce the burden on caregivers. Current assistive mobility platforms (e.g., robotic wheelchairs) often fail to incorporate user preferences and control, leading to reduced trust and efficiency. Existing shared control algorithms do not allow the incorporation of the user control preferences inside the navigation framework or the path planning algorithm. In addition, existing dynamic local planner algorithms for robotic wheelchairs do not take into account the social spaces of people, potentially leading such platforms to infringe upon these areas and cause discomfort. To address these concerns, this work introduces a novel socially-aware shared autonomy-based navigation system for assistive mobile robotic platforms.

Our navigation framework comprises a Global Planner and a Local Planner. To implement the Global Planner, the proposed approach introduces a novel User Preference Field (UPF) theory within its global planning framework, explicitly acknowledging user preferences to adeptly navigate away from congested areas. For the Local Planner, we propose a Socially-aware Shared Control-based Model Predictive Control with Dynamic Control Barrier Function (SS-MPC-DCBF) to adjust movements in real-time, integrating user preferences for safer, more autonomous navigation. Evaluation results show that our Global Planner aligns closely with user preferences compared to baselines, and our Local Planner demonstrates enhanced safety and efficiency in dynamic and static scenarios. This integrated approach fosters trust and autonomy, crucial for the acceptance of assistive mobility technologies in the built environment.

**BACKGROUND AND CHALLENGE**

Lack of access to independent end-to-end (E2E) mobility has life-altering implications for Persons with Disabilities (PWD). Mobility constraints in PWD reduce access to education, healthcare, and employment; exacerbate physical and mental stress leading to other chronic morbidities; and impose a significant burden on families, caregivers, the healthcare system, and society (Mortenson et al. 2005; Fehr et al. 2000). In the United States alone, nearly 2.7 million individuals experiencing mobility impairments rely on wheelchairs for their daily activities (Koontz et al. 2015). This number is expected to grow 7% annually (Flagg 2009) due to significant improvements in post-trauma survivorship (with some physical impairments) and longevity (Bureau 2016; King et al. 2013). Consequently, this growing necessity underscores the urgent need for the development and implementation of innovative assistive technologies that implement efficient systems for navigation and collision-avoidance, and can help PWD improve their mobility in the built environment. Assistive mobile robots, designed to aid those with mobility impairments, have shown considerable promise in enhancing the quality of life for PWD (Fardeau et al. 2023). These robots offer the potential for increased independence in the built environment, including homes, workplaces, and public spaces.

A critical aspect of assistive mobile robots is their navigation system, which determines how effectively and safely a robot can move through and interact with its environment (Rubio et al. 2019). These systems not only enable the robot to maneuver safely but also to adapt to specific user preferences and interact within dynamic spaces. For instance, user preferences could include the robot maintaining a certain speed or going in a direction preferred by the user, which is particularly crucial for users with physical disabilities who may require wider clearances to be comfortable. Socially aware navigation, on the other hand, involves the robot recognizing and respecting human social norms and personal spaces, such as not coming too close to people or appropriately navigating around groups without causing disruption. A properly implemented navigation system allows assistive robots to understand and adapt to the user's preferences, navigate efficiently in crowded or dynamic spaces, and interact with the environment in a socially acceptable manner (Gao and Huang 2022).

The navigation system can be divided into a Global Planner and a Local Planner. The Global Planner is responsible for generating an optimal and feasible path from start to destination, taking into account the user's specific preferences for the avoidance of crowded areas. This planner ensures that the overall path aligns with the user's needs and comfort. On the other hand, The Local Planner handles real-time navigation by adapting to immediate obstacles and environmental changes, incorporating social awareness to sensitively navigate human environments and maintain appropriate distances from people. It also allows users to manually adjust the robot's motion to meet their immediate needs and preferences. As shown in Fig. 1, the Global Planner combines the user's preferences to generate an optimal and feasible path, while the Local Planner integrates social awareness to enable the robot to anticipate and navigate human environments more gracefully.

The integration of the user's control and preference is crucial for fostering trust between the user and the assistive wheelchair, enhancing the user's willingness to rely on the assistive wheelchair for daily activities. In the absence of such features, users may experience significant challenges and reluctance, as the lack of synchronization between user needs and the wheelchair's operations can lead to a diminished sense of autonomy and security. Moreover, by integrating social awareness, an assistive wheelchair can anticipate and navigate human environments more gracefully, reducing the cognitive burden on users and enabling smoother, more natural interactions within social settings (Avelino et al. 2021). Without this capability, there are inherent risks that users may find

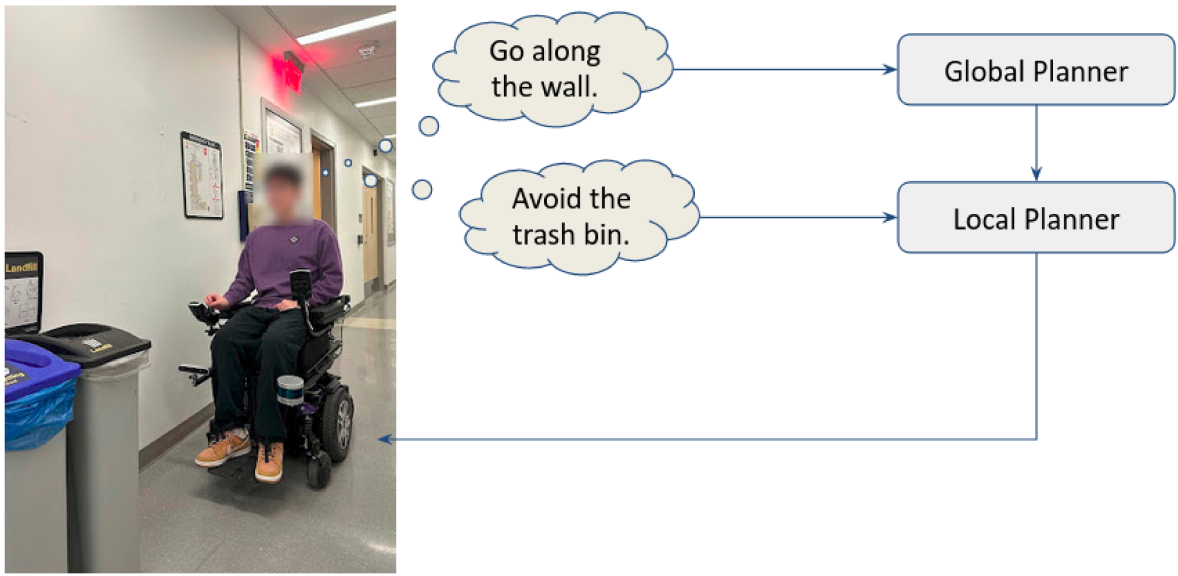

**Fig. 1.** Illustration of the shared autonomy-based navigation

robotic wheelchairs to be cumbersome and alienating, hindering their acceptance and utilization.

**Global Path Planning**

For the Global Planner, one of the main focus areas is to integrate the user's preferences into its workflow (Mokhtari et al. 2009). In order to increase the comfort and trust of the user, it is important for the Global Planner in the navigation system to be aware of the user's preferences. By aligning the robot's navigation strategies with these personal preferences, the system not only improves the user experience but also fosters a greater sense of reliability and safety, which are essential for user acceptance and satisfaction with autonomous systems. However, some of the navigation methods in the literature focus on modeling users' preferences in choosing different paths without allowing them to participate directly in the actual path selection or allowing them to control the whole navigation process. Although these autonomous navigation approaches can generate an optimal and smooth path, users cannot directly influence the navigation direction.

For example, (Chang et al. 2017) uses historical data to predict that a user prefers less crowded routes without allowing the user to override this choice in real time. This approach contrasts with systems that enable users to actively select their preferred route or adjust navigation settings during transit, thus providing a more interactive and personalized navigation experience. The direct involvement can significantly enhance user trust and satisfaction by making the navigation process more transparent and responsive to immediate user needs. Inspired by (Lu et al. 2014), a User Preference Field (UPF) layer is applied to a multi-layer cost map for global path planning. This proposed UPF map layer can successfully incorporate the user's preferences into path selection and generate collision-free paths that the user prefers in the environment.

**Local Path Planning**

Since the assistive wheelchair is designed to serve users, shared control plays an important role in gaining people's trust (Oksanen et al. 2020) and adapting itself into society (Yin et al. 2021; Cubuktepe et al. 2023; Desai et al. 2009). Traditional navigation systems either consider a fixed degree of human's control command velocity as the primary input to operate the robot or make a mobile robot navigate autonomously without a human's control. These approaches fail to consider the varying preferences of users, who may desire different levels of involvement in controlling the robot, thereby affecting their satisfaction and the usability of the technology. Some navigation methods, such as the Artificial Potential Field (APF) (Petry et al. 2010), allow users a fixed degree of control over the navigation process. While this approach considers user input, it does not account for the varying degrees to which individuals may wish to control the navigation, potentially limiting personalization and user satisfaction.

Furthermore, as users often are unfamiliar with the kinematics of a robotic platform (e.g., wheelchair), it is challenging for them to manually control the device to avoid all static and dynamic obstacles in the environment. This unfamiliarity makes it difficult for users to manually avoid obstacles, leading to potential collisions when mistakes occur in control. The current state of the art does not adequately address these user-centric concerns, emphasizing the need for more adaptable and intuitive control systems that can cater to the specific needs and skill levels of diverse users. This adaptability is essential for enhancing user experience and ensuring safe interaction with the robot in complex environments.

To address this issue, this paper proposes a Socially-aware Shared control-based Model

Predictive Control with Dynamic Control Barrier Function (SS-MPC-DCBF) that enables a Local Planner to let the user adeptly take control over the obstacle avoidance problem. The proposed method utilizes Dynamic Control Barrier Function (DCBF) to let the robot avoid the obstacles in advance and generate a safe collision-free trajectory (Jian et al. 2022). Additionally, a user velocity prediction-based social area detection is integrated into the Local Planner to make the robot avoid pedestrians' personal space/area and adapt to the social environment. Finally, a user can make their choice to take partial or complete control over the navigation process. This Local Planner not only maintains a socially respectful distance from ambient humans but also facilitates a more trustful and comfortable interaction for individuals controlling the robot, such as PWD navigating a wheelchair. The integration of advanced control mechanisms and social awareness presents a significant advancement in the development of user-centric, socially aware robotic systems.

**Research Objectives and Contributions**

Drawing upon the insights presented in (Petry et al. 2010; Jian et al. 2022; Wang et al. 2022), we propose a socially aware shared control-based dynamic navigation system that takes the user's preference in path choosing and the user's desire for control during the navigation process into consideration. As shown in Fig 2, we divide our navigation system into two parts: Global Planner used for global path choosing and Local Planner for executing the path plan and avoiding any collisions with dynamic obstacles. Generally speaking, our shared control-based navigation system desires to address four categories of challenges in built environments.

First, toward the issue that a user cannot directly choose their own preference-based path, our Global Planner takes the adaptive user's path preference into consideration in the global path planning algorithm and makes the navigation system generate a user desired collision-free path. For example, if a user prefers quieter, less trafficked routes, the Global Planner can adapt to prioritize these preferences in its path planning. This capability significantly enhances the navigation system's responsiveness to individual needs, improving user satisfaction and safety.

Second, in order to let the robotic platform adapt to the social environment, our approach integrates social areas (spaces) into the obstacle modeling pipeline in the Local Planner to prevent the robot from getting too close to any pedestrians or obstacles detected. This integration ensures the robot maintains a respectful distance from pedestrians and avoids encroaching on personal spaces, thereby preventing discomfort or safety issues.

Third, to create a safe and efficient path in a dynamic environment and avoid any future collision with moving obstacles, we leverage the Dynamic Control Barrier Function (Ames et al. 2019) to let the robot avoid the obstacles in advance, even when they are beyond a pre-defined proximity threshold. This ensures a smoother and safer navigation experience, maintaining an efficient path without sudden deviations that could confuse or alarm the user. Finally, to address the challenge of user distrust in autonomous navigation, our system employs shared control-based Model Predictive Control (MPC) as the Local Planner, allowing users to dictate their level of involvement in the navigation process.

This feature empowers users to either take direct control or let the robot navigate autonomously based on their immediate preferences and situations, enhancing their sense of security and control over the robot. Without these features, users might experience increased anxiety due to close proximity to other pedestrians, potential collisions with unforeseen obstacles, and a feeling of disconnection or lack of control over the robotic platform, ultimately leading to diminished trust and satisfaction with the technology.

Our proposed navigation approach has the following contributions:

- We propose a novel shared control-based navigation framework for assistive mobile robots integrating a user's desire for path preference and control during navigation.
- A Global Planner, designed based on user preferences, is developed and incorporated into the global planning pipeline, allowing users to customize their preferred global path.
- We propose the SS-MPC-DCBF Local Planner to generate safe collision-free trajectories in a dynamic environment.
- We integrate a detection and group space modeling method based on the personal and group spaces inside the Local Planner, which prevents the robot from intruding into other pedestrians' social spaces/areas.

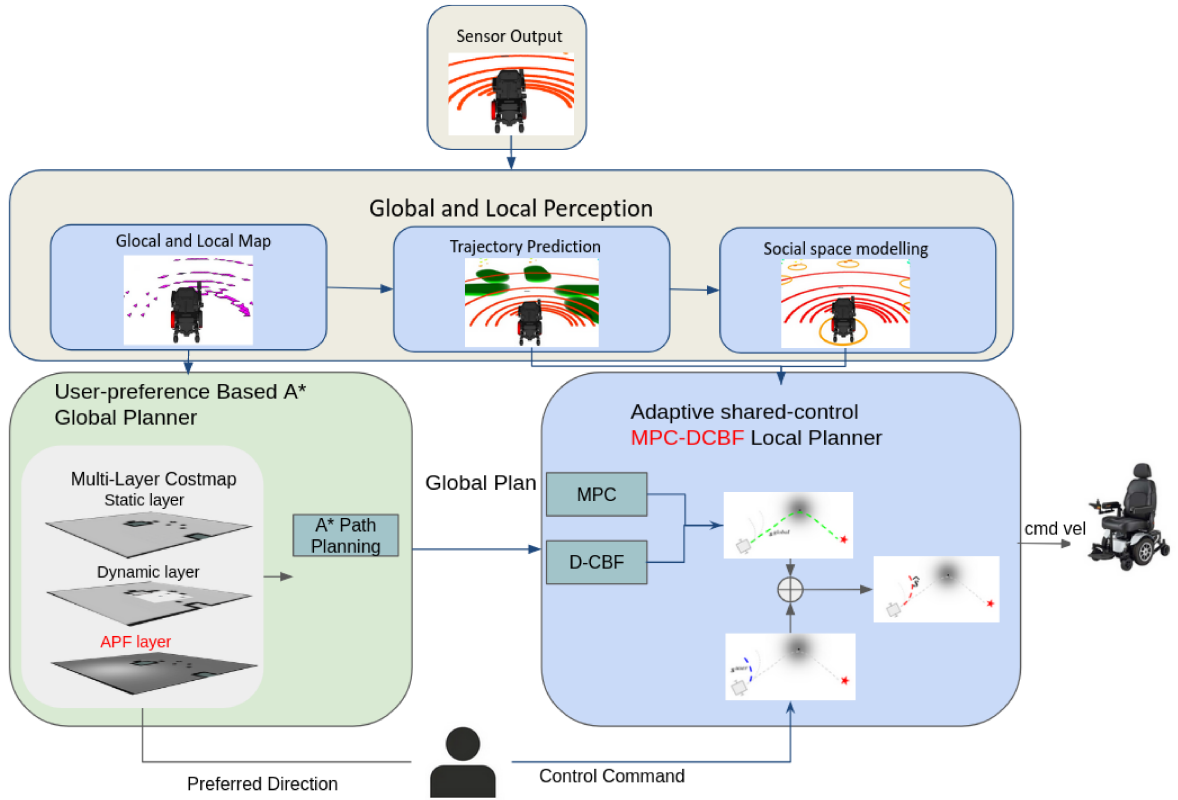

**Fig. 2.** The structure plot of the shared autonomy-based navigation framework.

- We designed different static and dynamic scenarios to test the adherence to user preferences of the Global Planner and the safety and efficiency of the Local Planner. The results show our proposed Global Planner exhibits greater alignment with user preferences compared to those generated by benchmark methods. In addition, our proposed Local Planner is shown to smoothly and robustly avoid dynamic obstacles exceeding other baseline methods.

This paper is organized as follows: Section 2 discusses a brief history of shared control-based navigation and the related technologies. Section 3 introduces our proposed shared autonomy-based navigation framework. Section 4 shows the evaluation results of the proposed Global and Local Planner, respectively. We conclude with our results and discuss the limitations and expected future work in Section 5.

**RELATED WORK**

In this paper, our goal is to create a robotic wheelchair that is capable of integrating human preferences (Shared Control-based Navigation), responding quickly to dynamic objects (Local Planner for Dynamic Obstacle Avoidance), and adapting to complex social environments (Socially-aware Navigation), which are the three specific domains this research addresses.

**Shared Control-based Navigation**

Considering a wheelchair's role as an intimately user-integrated robotic system, it becomes essential to incorporate user preference for global and local planning. Therefore, within our current framework, we have incorporated methods and concepts of shared control. This integration underscores the critical balance between autonomous robotic function and the adaptation to individual user preferences, a cornerstone in advancing user-centric robotic mobility solutions.

Most of the research in mobile assistive robotics extensively explored two ends of the spectrum - fully autonomous and fully manual. In the fully autonomous navigation field, a significant number of path planning and obstacle avoidance approaches are proposed. Potential field methods (Khatib 1986), A* graph search (Goyal and Nagla 2014), D* for dynamic environments (Stentz 2011), and Rapidly-exploring Random Tree (RRT) (Karaman and Frazzoli 2011) and others are among those widely studied. Recently, studies considering human presence as an interactive component requiring communication have brought a new dimension to the field of mobile robot navigation. (Chang et al. 2017) proposed iterative A* for the user to choose their preferred path and utilize Support Vector Machine (SVM) to allow the robot to imitate the users' preference. Although some autonomous navigation can generate an optimal path both globally and locally, the desire to feel safe and comfortable is not taken into consideration, which can cause human-robot trust issues when the robot navigates autonomously.

In the fully manual control field, a lot of work focuses on integrating people's preferences into the path planning pipeline and allowing the system to generate the user preference-based path. (Yanco 1998) proposed an assistive wheelchair system to provide different path and obstacle information on the interactive user interface (UI) to help the user make decisions on path planning.

Recently, (Mantha et al. 2020) utilized an indoor attribute-loaded graph network to take human preferences or physical constraints into account during path planning. Although a robotic assistive system in manual control can achieve high performance in global high-level planning, they are not effective with micro-level local planning and accuracy (Koontz et al. 2015).

Shared control-based navigation lies in the middle-ground that incorporates elements of both and has played an important role in gaining user trust during the robot navigation phase (Yin et al. 2021; Cubuktepe et al. 2023; Desai et al. 2009). Some research focuses on applications within the Global Planner, for instance, directly influencing global path planning through user input (Chang et al. 2017). However, such methods tend to be relatively inefficient in practical operation and may cause difficulties in human interaction. There are also studies proposing interfaces that allow users to add obstacles to the global map (Karanam et al. 2022), but this approach might lead to the problem of being unable to generate effective paths. Additionally, some prior studies have focused on applications in the Local Planner. For example, research based on the Artificial Potential Field (APF) (Khatib 1986) method maps the user's joystick operations as a force exerted on the robot (Petry et al. 2010). However, this method carries the risk of the robot being "frozen" in certain positions due to balanced forces.

In our work, the concept of shared control is implemented through enhancements for both global and local planners. We first integrated human preferences into the global planner by improving the A* algorithm. Subsequently, we incorporated these preferences into the local planner by refining the cost function of the Model Predictive Control (MPC).

**Local Planner for Dynamic Obstacle Avoidance**

In robotics and autonomous vehicle navigation, a "Local Planner" is part of the motion planning system that calculates short-term paths or maneuvers for the robot or vehicle. Regarding the Local Planner, currently there are many solutions available in the literature. For example, the Dynamic Window Approach (DWA) (Fox 1997) focuses on generating multiple paths from the robot's current state and then selecting the optimal one, considering both the direction of the robot's goal and obstacle avoidance. The Timed Elastic Band (TEB) (Verlag 2012) visualizes the planned path as an elastic band that gets stretched or compressed based on dynamic constraints, effectively molding the path according to the environment's constraints. The Artificial Potential Field (APF) (Khatib 1986) method conceptualizes the robot's movement as being influenced by artificial forces: attractive forces pulling it towards the goal and repulsive forces pushing it away from obstacles, navigating the robot through a balance of these forces. However, their performance can be significantly impacted in complex environments, such as those with many pedestrians (Kobayashi et al. 2021).

MPC (Model Predictive Control) is another widely used method in Local Planners, but most MPC implementations focus on lane tracking and static obstacle avoidance (Rösmann 2020; Turri et al. 2013) without further addressing obstacle avoidance for dynamic objects. To address the dynamic obstacle avoidance and to enhance the robot's obstacle avoidance capability, some algorithms that integrate the predicted trajectories of moving objects into the MPC's cost function can greatly enhance dynamic obstacle avoidance capabilities (Poddar et al. 2023; Wang et al. 2022). These approaches work by including the future positions of pedestrians as part of the MPC cost function, allowing the robot to anticipate and make decisions in advance to avoid pedestrians. However, a drawback of this approach is that MPC will not confine the optimization problem if the distance between the robot and obstacles is larger than a defined social distance.

Consequently, the robot only reacts when it is within a specified distance from obstacles. However, in complex and dynamic environments, this fixed distance may not be sufficient for the robot to respond and avoid obstacles effectively.

To mitigate this, some researchers have added Control Barrier Function (CBF) (Ames et al. 2019) constraints on top of MPC to further improve safety (Zeng et al. 2021a; Zeng et al. 2021b; Jian et al. 2022). CBF can be considered a constraint that keeps the robot away from obstacles. By incorporating CBF as a hard constraint into the MPC's cost function, the robot can make decisions about obstacles earlier rather than being solely restricted by the prediction horizon. This integration enables proactive obstacle avoidance, as the CBF constraint mathematically guarantees safe distancing from obstacles by dynamically adjusting the robot's trajectory in anticipation of future positions, thereby enhancing safety and efficiency in navigation. However, the original papers that proposed this approach simply model all obstacles as ellipses. Modeling humans and all other objects as ellipses overlooks the complexity of human behavior and psychology, especially considering that wheelchairs, in their daily use, need to interact extensively with people and uniform elliptical shapes may not always be the best projection of the users' dynamically-changing personal spaces. Therefore, we introduce social space and human motion prediction from social navigation to further optimize the navigation approach.

**Socially-aware Navigation**

Social navigation refers to the strategies and methods employed by robots to navigate through environments that involve interaction with humans or other agents, focusing on maintaining safe, socially acceptable, and efficient movement. It is introduced in the survey (Mavrogiannis 2023) that it is a vast field, and there have been many related studies in the past thirty years. These studies have evolved from initially focusing on simple engineering principles to now integrating multiple fields, including human-robot interaction, psychology, and sociology. Here, we focus on one specific issue: how to integrate social characteristics, or human information, into the planning process to achieve better results. For example, by incorporating data such as the positions, speeds, and even postures and facial expressions of pedestrians, the robotic platform can more effectively plan its avoidance paths and interactions with pedestrians.

One of the most common and direct methods to incorporate human information is to predict human behaviors (mainly motion trajectories) (Alahi 2016; Gupta 2018) and then integrate this predictive information into planning. Examples of this approach include incorporating pedestrian trajectory prediction into the cost function of an MPC framework (Wang et al. 2022; Poddar et al. 2023). Some approaches are modeling human behavior and trajectories from a geometric or topological representation perspective before integrating them into the planner (Mavrogiannis 2017; Mavrogiannis 2020). Others, starting from social norms and human behavioral habits, introduce social and psychological knowledge, such as the concept of social space described in their studies (Hayduk 1994; Poddar et al. 2023).

Social space is a concept that pertains to the invisible buffer zone individuals naturally maintain around themselves to feel comfortable and unthreatened while moving in their environment. Discovered through extensive observation and experimentation, it reveals that individuals have a relatively comfortable personal space while walking (Kirby 2010). Based on social space, the concept of social groups has been utilized to enable robots to avoid obstacles in dense pedestrian environments (Gadol 1981; Wang 2020; Wang et al. 2022). The concept of social groups emerges from substantial observation and experimentation to account for the collective movement and interaction patterns of these groups, indicating that people often walk in groups. There are also studies not focused on integrating specific human behavioral information but instead concentrating on learning the principles behind human behavior to enhance robot decision-making (Yu Fan Chen; Liu 2016).

In this paper, to enhance the navigational performance of robotic wheelchairs in complex social environments, we have incorporated some concepts from social navigation. These include the prediction of pedestrian trajectories, modeling of pedestrian social spaces, and ultimately integrating this information into the MPC-DCBF framework so that our algorithm can achieve safety and efficiency in navigation within complex environments, especially in pedestrian settings.

**RESEARCH METHODOLOGY**

As depicted in Fig. 2, the navigation framework comprises three main components: social area detection, a user-preference-driven global planner, and an SS-MPC-DCBF local planner. Fig. 3 illustrates the comprehensive system flowchart. The Global Planner utilizes an A* algorithm (Goyal and Nagla 2014), tailored to user preferences, which processes the occupancy grid map (OGM) alongside the user's preferred direction, thereby generating a user-preference field (UPF) layer. Here, the constructed OGM and articulated user preferences feed into the global planning module to generate an initial navigational strategy. This global plan, along with inputs from 3D LiDAR sensors and user control commands, is further processed through our Local Planner. The Local Planner employs the MPC-DCBF algorithm that incorporates a shared control mechanism to output the final command velocities for the robot's movement. This design enables users to assert control over both global and local planning phases, thereby enhancing overall control accuracy and responsiveness.

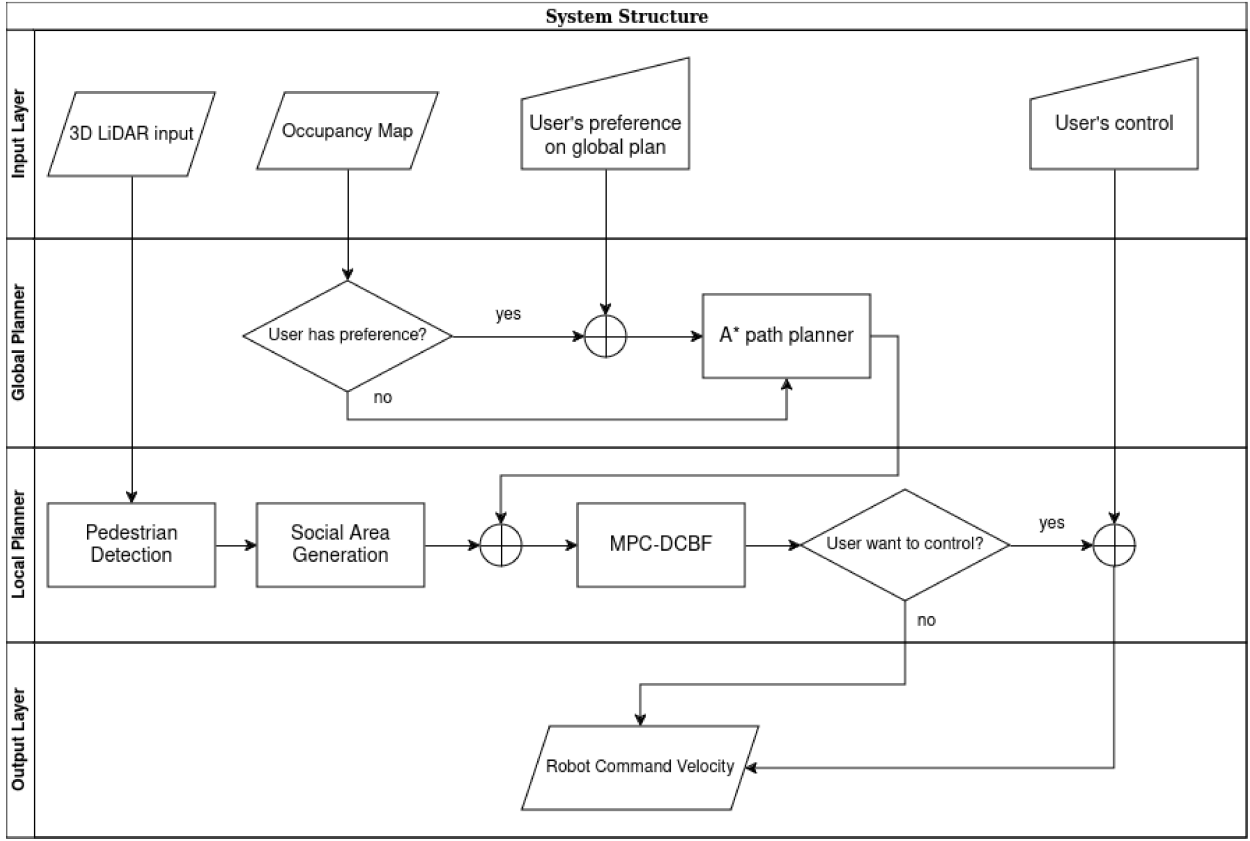

**Fig. 3.** The flow chart of the shared autonomy-based navigation framework.

**User preference-based Global Planner**

We model our global planner into multi-layer costmap construction and A* path planning. The workflow of the global planner is shown in Fig. 4. In the framework, the user can choose to include their preference in the generation of the global plan. The user preference field generation is discussed below.

Inspired by Gaussian Random Field theory (GRF) (Clifford 1994), we create UPF, which is a 2D GRF around the preference point that is chosen by the user, which is shown in Fig. 5. The multi-layer costmap is then fed into A* path planning pipeline and generates the global path. In the UPF layer, we set the cost near the user preference point chosen by the user lower than others to create a "cost valley" which is expressed in (2). Because A* path planning will always try to find a path that minimizes the cost and the cost near the center of UPF is lower, the path generated will be attracted towards the center.

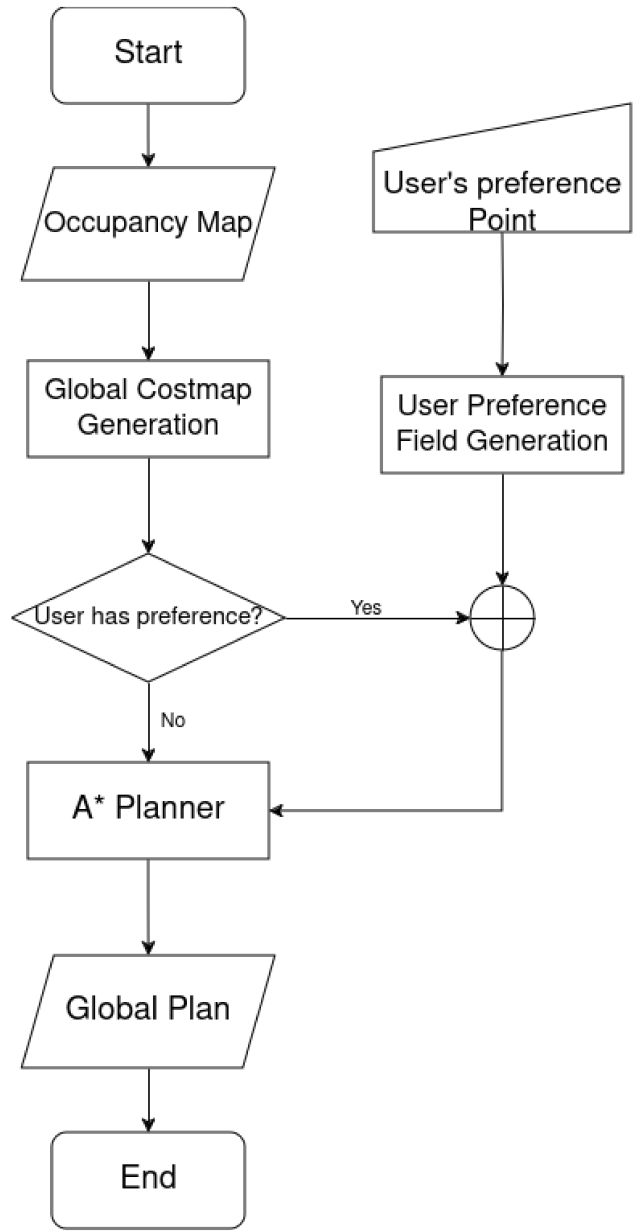

**Fig. 4.** The flow chart of the user preference-based global planner

The mean value of this 2D Gaussian distribution is located at the preference point. In order to ensure that this GRF will be able to successfully attract a robotic wheelchair to go through the user's preference direction, the standard deviation of the distribution, $\sigma$, is decided by the distance between the robot and the preference point $d_r$. To achieve the effect above, we set the range between the robot and the preference field to cover 90% of the distribution. The calculation of $\sigma$ is shown in (1).

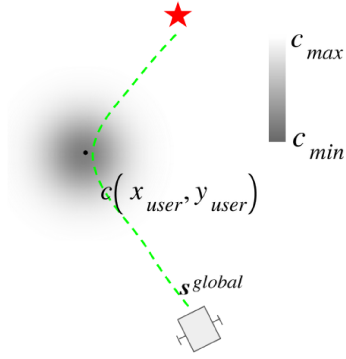

**Fig. 5.** The illustration of the user preference-based global planner

$$P(d_r) = \int_0^{d_r} \frac{z}{\sigma^2} e^{-z^2/2\sigma^2} dz = 1 - e^{-d_r^2/2\sigma^2}$$
$$1 - e^{-d_r^2/2\sigma^2} = 0.9 \quad (1)$$
$$\sigma = \frac{1}{\sqrt{-2 \ln 0.1}} d_r \approx 0.465 d_r$$

where $P(d_r)$ represents the probability within the circle range from the center to the robot location and $P(\cdot) \in [0, 1)$. In addition, according to (Lamprianidis 2018), we map the distribution value into the range where most user preferences should be expressed in the costmap. The equation is shown in (2).

$$d = \sqrt{(x_{user} - x)^2 + (y_{user} - y)^2}$$
$$p(x, y) = \frac{d}{\sigma^2} e^{-d^2/2\sigma^2} \quad (2)$$
$$c(x, y) = c_{max} - \frac{(p_{min} - p(x, y))(c_{max} - c_{min})}{p_{min} - p_{max}}$$

In the equation above, $(x, y)$ represents the location of a pixel in the map, $(x_{user}, y_{user})$ represents location of the preference point, $d$ represents the distance between the pixel and the user preference point, $p_{max}$ and $p_{min}$ is the maximum and minimum probability density value within the whole

**Algorithm 1** User-preference field

**Input:** $x_{user}, y_{user}, costmap_0, N_x, N_y, \Sigma_{user}$
**Output:** $costmap_{user}$

1:   $costmap_{user} \leftarrow costmap_0$      ▷ Initialization
2:   **for** $x = 0$ to $N_x$ **do**
3:     **for** $y = 0$ to $N_y$ **do**
4:       $p(x, y) \sim \mathcal{N}((x_{user}, y_{user})), \Sigma_{user})$      ▷ Calculate the probability density
5:       $c(x, y) \leftarrow Convert2MapValue(p(x, y))$
6:       **if** $costmap_{user}(x, y) < c(x, y)$ **then**      ▷ Whether is occupied by static obstacles
7:         $costmap_{user}(x, y) \leftarrow c(x, y)$
8:       **end if**
9:     **end for**
10: **end for**
11: **return** $costmap_{user}$

**TABLE 1.** Variable Definition.

| Variable | Definition | Variable | Definition |
| --- | --- | --- | --- |
| $x_{user}$ | x position of the center for user preference point | $y_{user}$ | y position of the center for user preference point |
| $costmap_0$ | The original costmap | $costmap_{user}$ | The costmap after applying user-preference field |
| $N_x$ | The total length of the costmap | $N_y$ | The total width of the costmap |
| $\Sigma_{user}$ | The covariance of the UPF | $d$ | Distance between the pixel and the user preference point |
| $p(x, y)$ | The probability density of the pixel | $c(x, y)$ | The cost value of the pixel |

map, $c(x, y)$ represents the cost value in the map pixel and $p(x, y)$ represents the probability density of the Gaussian distribution. The algorithm for calculating the UPF is shown in Algorithm 1 and TABLE 1 shows the definition of the variables.

**SS-MPC-DCBF Local Planner**

Upon deriving a global path plan from the user preferences-based path planner, our methodology employs the SS-MPC-DCBF for generating the command velocity, which is crucial for robot navigation and ensuring a collision-free trajectory within dynamic social settings. As shown in Fig. 6, the 3D LiDAR input is used to detect pedestrians using DBSCAN and generate the social area based on the velocity detection. Then, if the user wants to control the wheelchair beyond the generation of the autonomous local path from MPC-DCBF, the user can adeptly add control to the final robot command velocity. The social area generation and shared control-based MPC-DCBF are discussed below.

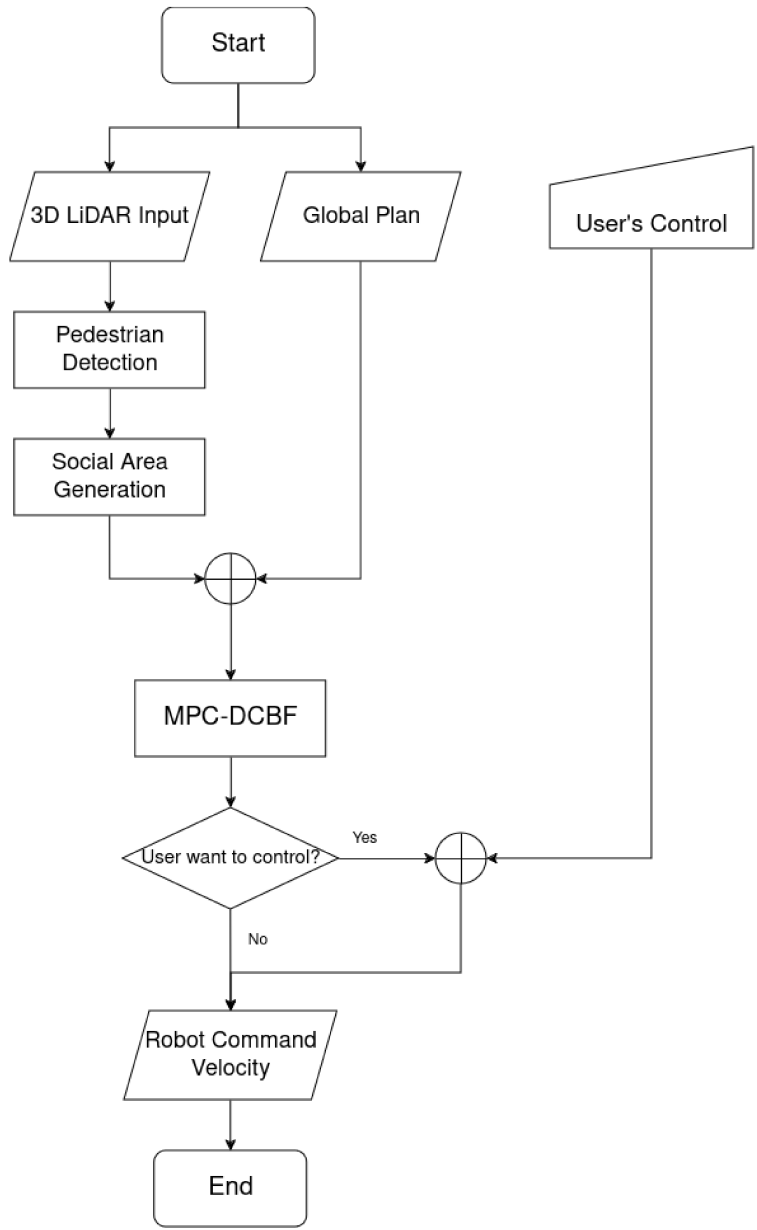

**Fig. 6.** The flow chart of SS-MPC-DCBF local planner

*Social Area Detection*

As shown in Fig. 7, the human detection and social area construction is divided into three steps: Human Detection, Kalman Filter-based Trajectory Prediction, and Social Area generation. First, for the human detection part, since the end-point will clustered if it is located at any person's body, we use Density-Based Spatial Clustering of Applications with Noise (DBSCAN) (Jiang et al. 2023) to cluster the endpoints combined set $\boldsymbol{q} = \bigcup_{i=1:n} q_i$ from 3D LiDAR data to detect pedestrians and use the average position of each end-point to represent the location of the people. The equation is shown below in (3):

$$\begin{aligned} \boldsymbol{P} &\longleftarrow DBSCAN(\boldsymbol{q}|\epsilon_d) \\ \bar{\boldsymbol{x}} &= \frac{1}{n}\sum_{i=1}^{n} \boldsymbol{x}_{q_i} \end{aligned} \quad (3)$$

where $\boldsymbol{P}$ represents the set of pedestrians detected and $\epsilon_d$ represents the distance threshold for point clustering. $\boldsymbol{q}$ represents the set of end-points that is used for detection. $\bar{\boldsymbol{x}} = (\bar{x}, \bar{y})$ is the position of the pedestrian detected from the average positions of each end-point $\boldsymbol{x}_{q_i} = (x_{q_i}, y_{q_i}), q_i \in \boldsymbol{q}$.

After the detection of pedestrians, we use Kalman Filter (KF) (Kalman 1960) to track each pedestrian and make velocity prediction based on the previous positions based on the previous positions and velocity. The equations for KF prediction are shown in (4):

$$\begin{aligned} \boldsymbol{x}_{k+1|k+1}^{p_i} &= KF(\boldsymbol{x}_{k|k}^{p_i}, P_{k|k}^{p_i}) \\ \boldsymbol{v}_{k+1}^{p_i} &= \frac{\boldsymbol{x}_{k+1|k+1}^{p_i} - \boldsymbol{x}_{k|k}^{p_i}}{T} \\ p_i &\in \boldsymbol{P} \end{aligned} \quad (4)$$

where $\boldsymbol{x}_{k+1|k+1}^{p_i}$ is the future state predicted from people's current state $\boldsymbol{x}_{k|k}^{p_i}$ and error covariance $P_{k|k}^{p_i}$. The velocity $\boldsymbol{v}_{k+1}^{p_i}$ is calculated from the two conjoint states with sample time $T$.

Inspired by (Kirby 2010), the personal space can be modeled as an Asymmetric Gaussian function (AGF). This function is used in the literature to define the personal space decided by a person's velocity. Since individuals are usually more aware of people in front of them than the orthogonal sides, as shown in Fig. 8, we construct the shape of the social area as an "*Egg*"shape instead of a circle around the center positions of each pedestrian. The AGF is separated into two front and rear distributions due to velocity. These two distributions share the same covariance in the perpendicular direction to the velocity but have a different covariance value in the parallel direction. The equation for generating the social distance is shown below in (5):

$$\begin{aligned}
\sigma_h &= \max(\frac{1}{2}, 2|\boldsymbol{v}_k^{p_i}|) \\
\sigma_s &= \frac{2}{3}\sigma_h \\
\sigma_r &= \frac{1}{2}\sigma_h \\
\sigma &= \begin{cases} \sigma_h & \text{for } 0 < \delta_k^{p_i} < \pi \\ \sigma_r & \text{for } \delta_k^{p_i} \geq \pi \text{ or } \delta_k^{p_i} \leq 0 \end{cases} \\
l_k^{p_i} &= \frac{c}{\frac{\cos^2(\delta_k^{p_i})}{2\sigma^2} + \frac{\sin^2(\delta_k^{p_i})}{2\sigma_s^2}}
\end{aligned} \quad (5)$$

where $\sigma_h$ represents the covariance in front direction same direction whereas $\sigma_r$ represents the covariance in opposite direction. $\sigma_s$ represents the covariance in the side direction. $l_k^{p_i}$ represent the distance in the direction which has $\delta_k^{p_i}$ angle with the moving direction of pedestrians. $c$ is a scale factor.

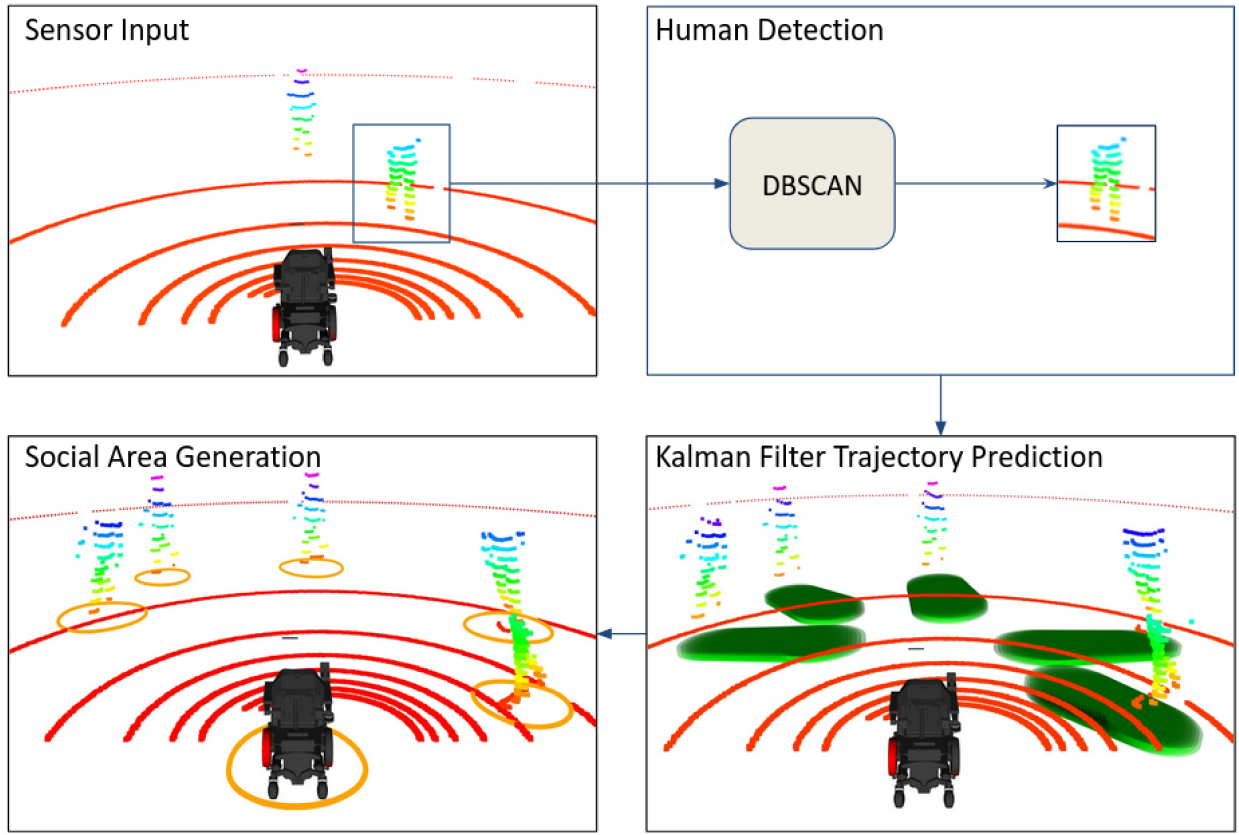

**Fig. 7.** Pedestrain detection and social area generation

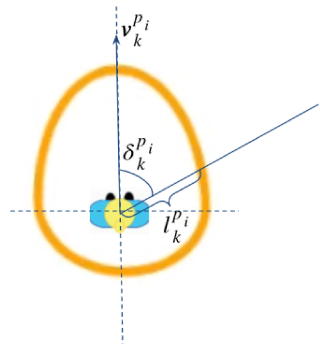

**Fig. 8.** Illustration for social area construction

In summary, our method starts by detecting people and predicting their velocity. With this information, we create a social area around each person for the robot to recognize. The following Local Planner treats these social areas as obstacles, which helps the robot avoid bumping into people, making its movement safe and respectful of personal space in places where people are moving around.

*Shared control based socially aware MPC-DCBF*

It has been proven that Dynamic Control Barrier Functions (DCBF), which consider the movement of obstacles, can be used for unmanned aerial vehicles (UAVs) and mobile robots (Zeng et al. 2021b; Glotfelter et al. 2019; Singletary et al. 2020; Ames et al. 2019). However, the current MPC-DCBF Local Planner shapes the obstacles as an ellipsoid based on the clustering result of DBSCAN and does not take the user's control into the path planning pipeline. This can lead to two issues: First, since the ellipsoid generated from the clustering result is too close to the body of the person, the robot will be very likely to generate a local path that intrudes into the personal area of other pedestrians. Second, the current MPC-DCBF method does not take the user's control into the local planning pipeline. Since the sensor can have blind spots and the robot can encounter unexpected and extremely unsafe situations, the MPC-DCBF Local Planner can generate a risky path that can possibly cause collisions with pedestrians.

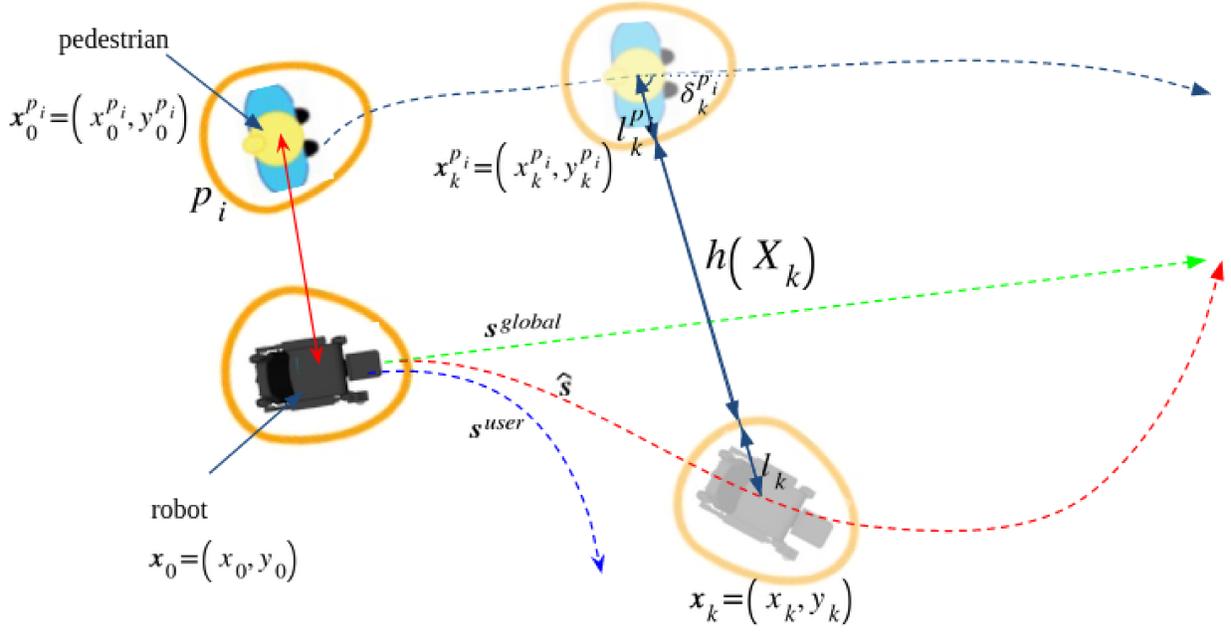

**Fig. 9.** Illustration of SS-MPC-DCBF

To address these two problems, inspired by (Jian et al. 2022), we proposed SS-MPC-DCBF, which takes the pedestrian's social area and user's control into the local planning pipeline, which will help the robot avoid the obstacles and generate a safer and more effective path for the robot. SS-MPC-DCBF is introduced below.

At time step $k$, we define robot position, pedestrians position as $\boldsymbol{x}_k = [x_k, y_k, \theta_k]^T \in \mathbb{R}^3$ and $\boldsymbol{x}_k^{p_i} = [x_k^{p_i}, y_k^{p_i}, \theta_k^{p_i}]^T \in \mathbb{R}^3$ respectively. The social area of robot and pedestrians are defined as $O_k = [\sigma_h, \sigma_s, \sigma_r]^T \in \mathbb{R}^3$ and $O_k^{p_i} = [\sigma_h^{p_i}, \sigma_s^{p_i}, \sigma_r^{p_i}]^T \in \mathbb{R}^3$. The set of all obstacle positions and shapes is described below in (6):

$$O_k^p = \bigcup_{i=1,2,\ldots,n} O_k^{p_i}$$
$$\boldsymbol{x}_k^p = \bigcup_{i=1,2,\ldots,n} \boldsymbol{x}_k^{p_i} \tag{6}$$

We combine all obstacle positions and shapes together and define the state variable as $X_k = [\boldsymbol{x}_k, \boldsymbol{x}_k^p, O_k, O_k^p] \in \mathbb{R}^{3\times(2n+2)}$. For safe-critical control, we define a safe set $C$ as the super set of the control barrier function $h$ as shown in (7):

$$C = \{X_k \in \mathcal{X} : h(X_k) \geq 0\}, \mathcal{X} \in \mathbb{R}^{3\times(2n+2)} \tag{7}$$

As shown in Fig. 9, we define the Control Barrier Function $h(X_k)$ from the distance between the robot and pedestrians and their social area in a quadratic form. The equation is described below in (8):

$$h(X_k) = \|\boldsymbol{x}_k - \boldsymbol{x}_k^{p_i}\|_2 - l_k - l_k^{p_i} \; for \; k = 0, 1, \ldots, N - 1 \tag{8}$$

where $\boldsymbol{x}_k$ and $\boldsymbol{x}_k^{p_i}$ represent the positions of robot and pedestrians around it. $l_k$ and $l_k^{p_i}$ are the distance values according to the social area generation shown in Fig. 9.

For our discrete-time system, we define the DCBF constraints as below in (9):

$$\Delta h(X_k, u_k) \geq -\gamma h(X_k), 0 < \gamma \leq 1 \tag{9}$$

where $\Delta h(X_k, u_k) := h(X_{k+1}) - h(X_k)$. $\gamma$ is a scalar factor controlling the effect of DCBF. The lower bound of $h(X_k)$ decreases exponentially with the rate $1 - \gamma$. After defining the constraints from the DCBF, we can formulate our shared control-based MPC problem. We employ a non-linear MPC formulation (Wang et al. 2022) for navigation in an indoor environment.

To formulate the MPC problem, we first describe the robot's dynamic model by a discrete-time equation $\boldsymbol{x}_{k+1} = f(\boldsymbol{x}_k, \boldsymbol{u}_k)$, where $\boldsymbol{x}_k = [x_k, y_k, \theta_k]^T \in \mathcal{X} \subset \mathbb{R}^3$ and $\boldsymbol{u}_k = [v_k, \omega_k]^T \in \mathcal{U} \subset \mathbb{R}^2$. According to (Filipescu et al. 2011), the kinematic equation of the differential robot is shown below:

$$\Delta x = \begin{bmatrix} \cos(\theta_k) & 0 \\ \sin(\theta_k) & 0 \\ 0 & 1 \end{bmatrix} u_k \Delta t \quad (10)$$

$$x_{k+1} = x_k + \Delta x$$

We then employ a non-linear MPC formulation (Filipescu et al. 2011) for navigation in an indoor environment. The optimization problem is shown below in (11):

$$\begin{aligned}
u_k^* &= \arg\min_{u_k \in \mathcal{U}} \sum_{k=0}^{N-1} \mathcal{J}_k(x_k, \hat{x}_k, u_k) + \mathcal{J}_N(x_N) \\
s.t.\ & x_{k+1} = f(x_k, u_k) \\
& x_k \in \mathcal{X} \\
& u_k \in \mathcal{U} \\
& \Delta h(X_k, u_k) \geq -\gamma h(X_k),\ 0 < \gamma \leq 1
\end{aligned} \quad (11)$$

where $\mathcal{J}_k$ represents the stage cost of each time step and $\mathcal{J}_N$ represents the terminal cost at the last time step. $f$ represents the kinematic dynamic of the differential robot model shown in (10). $\mathcal{X}$ and $\mathcal{U}$ are the feasible and reachable set of state and input. $x_k$ and $\hat{x}_k$ represent the actual state and reference state of the robot.

In order to let the user take partial control over the robot's motion, we take the predicted state generated by the user's control sequence into cost function $\mathcal{J}_k$ shown in (12).

$$\begin{aligned}
\mathcal{J}_k(x_k, \hat{x}_k, u_k) &= \mathcal{J}_k^s(x_k, \hat{x}_k) + \mathcal{J}_k^u(u_k) \\
\mathcal{J}_k^s(x_k, \hat{x}_k) &= (x_k - \hat{x}_k)^T Q_s (x_k - \hat{x}_k) \\
\hat{x}_k &= \eta(i) x_k^{user} + (1 - \eta(i)) x_k^{global} \\
\mathcal{J}_k^u &= u_k^T Q_u u_k
\end{aligned} \quad (12)$$

where $\mathcal{J}_k^s$ is the state cost and $\mathcal{J}_k^u$ is the input cost. $\hat{x}_k$ is the reference trajectory considering both the global path reference state $x_k^{global}$ and user's control sequence $x_k^{user}$. $\eta(i) \in [0, 1]$ is the weight of the user's control sequence, which will decide how much human control the robot will take. $i$ is the number of control signals that the user gives within a time frame. To calibrate user control levels in our system, we developed a weight function based on user control frequency, reflecting their intent to steer the wheelchair. We use an exponential function to model the user's intent shown in (13).

$$\eta(i) = 1 - e^i \tag{13}$$

As shown in Fig. 9, the reference path combines the user's control signal and the global path plan. The SS-MPC-DCBF Local Planner will then allow the robot to follow the combined path while avoiding intrusions into the present pedestrians' social area.

**EXPERIMENTAL RESULTS**

Our evaluation is divided into two parts - Global Planner evaluation and Local Planner evaluation. In this section, we first introduce our experimental setup for the Global Planner and the Local Planner. Then, for the Global Planner evaluation part, we evaluate our proposed user preference-based Global Planner by comparing it with the state-of-art user preference-based path planning method over time, trajectory length, and error to the ground truth user desired path. For the Local Planner part, we evaluate our SS-MPC-DCBF in both static and dynamic environments by comparing our algorithm with other state-of-the-art Local Planners.

**Experimental Setup**

We designed several different environments for testing the robot navigation system. Shown in Fig. 10, we separate our experiment scenes into dynamic experiments and static experiments. For the Global Planner evaluation, as shown in the right figure, we created an open area that simulates a large office environment where the user's preferences can be applied. In this open environment, we enable the robot to pass the central obstacle and avoid going from the right side, where most obstacles are located.

For the Local Planner, we conduct two different experiments to test its performance in both static and dynamic scenarios. In the static experiment, we created a narrow and challenging corridor environment where random pedestrians and obstacles are present, which is shown in the right figure. We compared the performance of this system with fully autonomous and manual control strategies. As for the dynamic experiment, we designed three experimental conditions involving robot navigation under different crowd behaviors that a robot could encounter in a crowded space, which will be discussed in the Local Planner evaluation section.

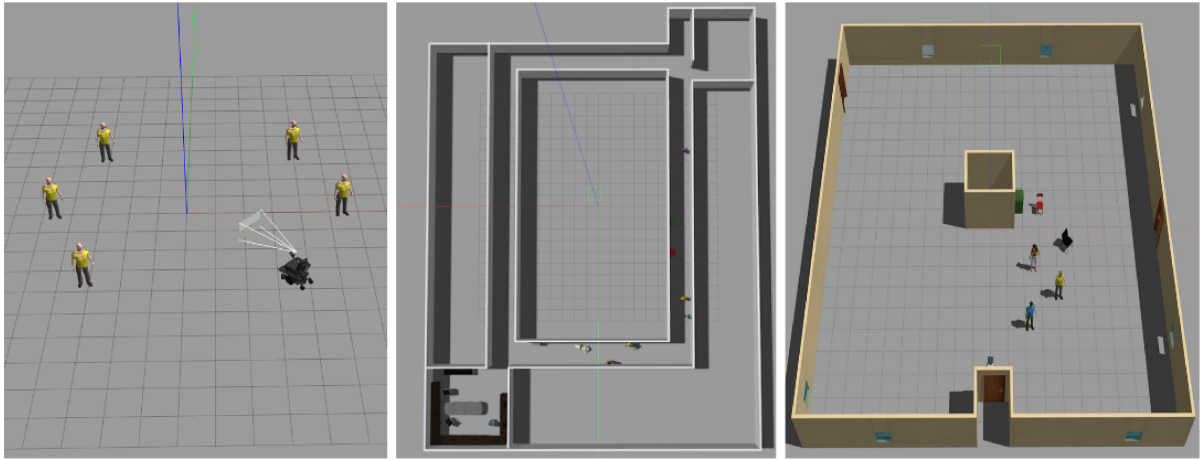

**Fig. 10.** Static and Dynamic simulation experiment

Regarding the simulated robot model (i.e., Digital Twin), we created a 3D simulated wheelchair model based on the widely used Quantum Q6 Edge 2.0 wheelchair. A LiDAR and a depth camera are integrated into the wheelchair, enabling it to map the environment and detect obstacles during indoor navigation. The LiDAR perception range is set to be $10m \times 10m$. The simulation model is shown in Fig. 11.

**Global Planner Evaluation**

To evaluate the efficacy of our novel global planning strategy, we conducted a series of experiments featuring a robotic wheelchair navigating through a semi-open environment. This environment was characterized by a dense array of obstacles on the right side, emulating an office setting. Despite similar distances on either side, the right-hand path presented numerous obstacles, invariably slowing the robot's progress.

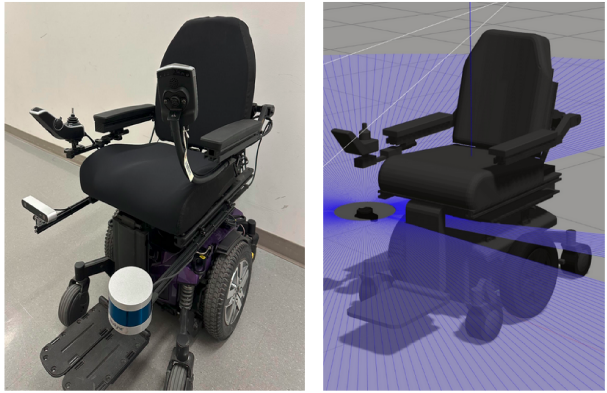

**Fig. 11.** Wheelchair model and Digital Twin in Gazebo

Our analysis compared the proposed UPF (User Preference Field)-based A* algorithm against the Iterative A* (Chang et al. 2017), a leading user preference-aware navigation technique cited in current research. Key performance metrics—namely, time to completion, trajectory length, and deviation from the user-preferred route—are used to evaluate our Global Planner's effectiveness and appeal in path selection. The adherence to the user's preferred path was posited as a critical metric, influencing the user's trust in the robot's navigational decisions and overall safety (Zhang et al. 2022).

The experiments are executed across ten varied scenarios, encompassing a range of obstacle configurations, to derive average values for time, trajectory length, and deviations from the intended path.

As shown in Fig. 12a, the qualitative result shows that the user-desired path(green) and the global paths generated by normal A*(blue), iterative A*(orange), and the proposed UPF-based A*(red). From the result, the error between the user-desired path and the proposed UPF-based path is the smallest compared with the other two.

As for the quantitative result, in Fig. 12b, The error-time plot of the proposed UPF-based A*, normal A*, and iterative A*. A higher error means the user desires the path less. Besides the error metrics, we also compared the time and trajectory length with the two baseline methods. The result is shown in TABLE 2. From the results, we observe that UPF-based A* has higher efficiency and can generate paths closest to the user's desired path.

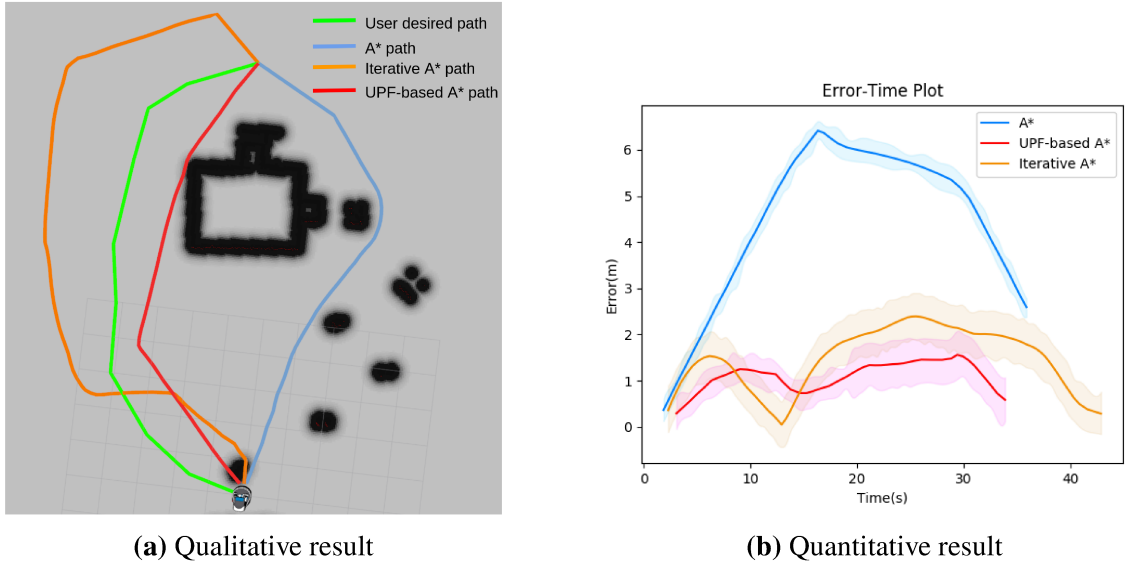

| | |
|:-:|:-:|
| **(a)** Qualitative result | **(b)** Quantitative result |

**Fig. 12.** The evaluation result of the global planner.

**TABLE 2.** Quantitative results of UPF-based A*.

| Method | Time | Trajectory Length | Error |
|:-:|:-:|:-:|:-:|
| UPF-based A* (Ours) | **33.90s** | **13.64m** | **1.10m** |
| Iterative A* (Chang et al. 2017) | 42.89s | 21.67m | 1.48m |
| A* (Goyal and Nagla 2014) | 35.90s | 14.32m | 4.14m |

The comparative analysis revealed that the UPF-based approach minimized deviations from the user's desired path more effectively than the alternatives. The standard A* algorithm frequently opted for paths through denser obstacle fields, while Iterative A*'s choices were constrained by the dimensions of virtually represented obstacles. Our UPF-based A* methodology allows users to influence the path selection directly by specifying the UPF center, thus ensuring the generated path closely mirrors the user's preferences. Fig. 12b, within a 15-second window, the UPF-based A* paths exhibit greater alignment with user preferences compared to those generated by benchmark methods.

**Local Planner Evaluation**

As for the Local Planner evaluation, we separate our algorithm into two different experiments, the static experiment and the dynamic experiment. For the indoor static experiment, we create a challenging and narrow corridor environment to test the performance of our SS-MPC-DCBF Local Planner. For the dynamic experiment, we designed three different crowded scenes for the robot to test its performance of obstacle avoidance and efficiency.

*Static Experiment*

As shown in the middle figure of Fig. 10, we create four corridor scenes where several obstacles are distributed around. This scene is used to test the performance of our SS-MPC-DCBF Local Planner compared with fully autonomous navigation and a fully manual control algorithm. After the experiment, we let the user rate their satisfaction based on a 1-5 scale. We assessed the planner based on time, success rate, collision rate, user satisfaction, trajectory length, and angle difference summary representing smoothness (Guillén Ruiz et al. 2020) against fully manual and autonomous controls. The results are presented in Fig. 13.

The findings from our study reveal that the proposed SS-MPC-DCBF framework significantly surpasses traditional fully manual and autonomous navigation modes in terms of efficiency and user satisfaction. Although manual control provides the potential for users to have the most direct routes, their lack of familiarity with the wheelchair's dynamics makes the robot hard to avoid static obstacles, increasing the risk of collisions and decreasing overall success rates. Furthermore, a lack of knowledge of the robot's dynamics often leads users to make excessive adjustments to its direction when navigating close to obstacles, leading to trajectories that are notably less smooth.

In contrast, while autonomous navigation is adept at identifying and following the shortest possible path and generally succeeds in avoiding obstacles, it faces difficulties in more complex scenarios, such as navigating around centrally located obstacles or dealing with sharp turns. In these instances, autonomous systems may require extended periods to generate a feasible route or, in certain cases, might cause the "freezing robot" problem (Trautman and Krause 2010).

Our proposed SS-MPC-DCBF Local Planner merges the strengths of both autonomous and manual navigation approaches. Under less demanding conditions, the autonomous mode enables the robot to proceed smoothly and securely. Conversely, in more complex scenarios, the system allows for user intervention, allowing partial control to navigate through challenging environments more effectively. This hybrid approach enhances navigation efficiency, offering a balanced solution that leverages the benefits of both control schemes.

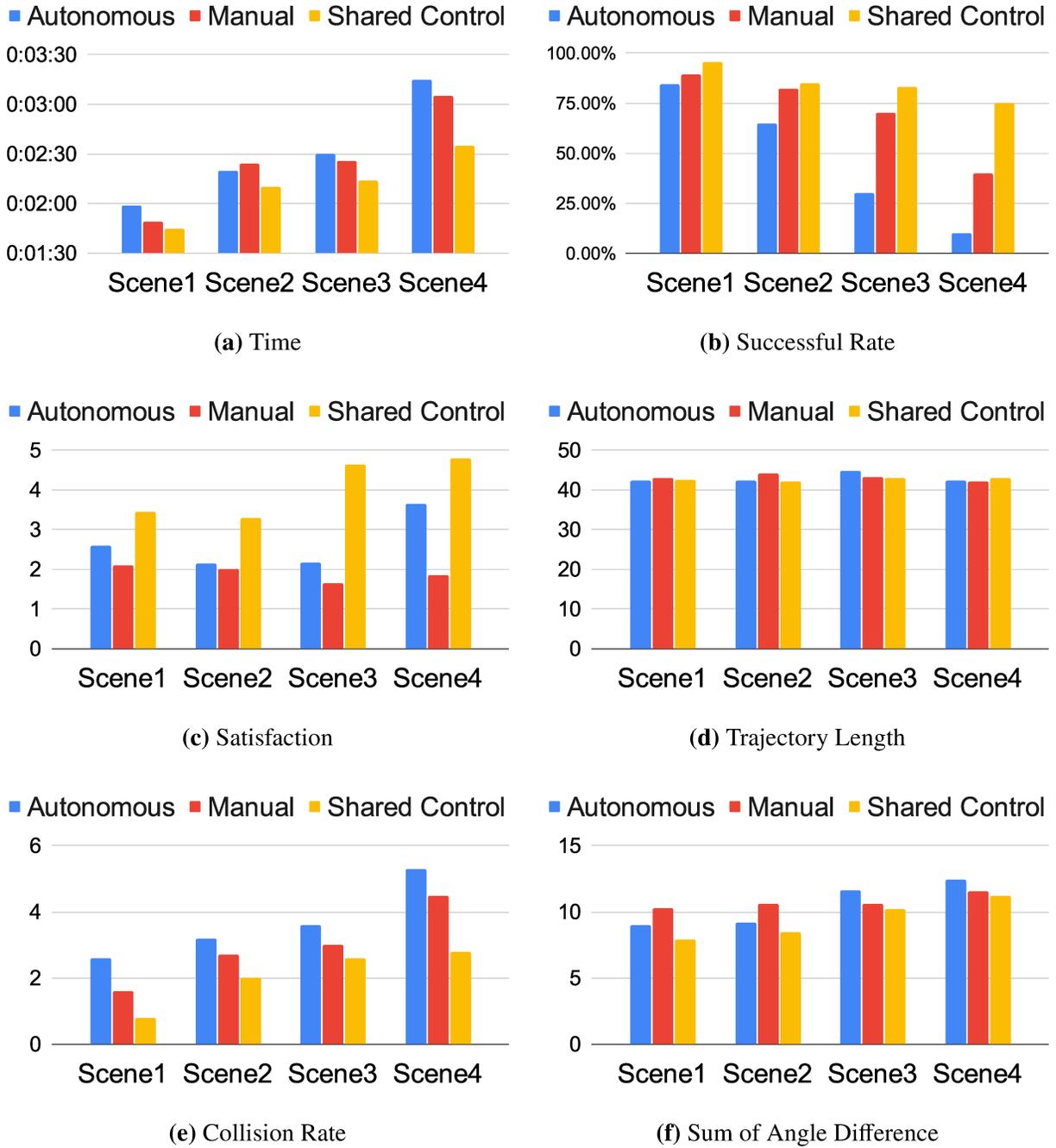

**Fig. 13.** The evaluation result of the proposed SS-MPC-DCBF local planner

*Dynamic Experiment*

According to (Poddar et al. 2023), we create three different dynamic scenarios that the robotic wheelchair may encounter in a crowded environment to test the performance under social constraints. The three scenes are shown in Fig. 14 and introduced below:

- **Crossing Experiment**: Five pedestrians and the robot move between the corners of the workspace. Pedestrians are instructed to navigate naturally with a normal walking speed. In this case, the users and robot are reactive to each other and are able to avoid each other. The setup is shown in Fig. 14a.
- **Aggressive Experiment**: One pedestrian and the robot switch each other's position. In this scenario, the robot will ignore the existence of the pedestrian and move to its goal "aggressively". The robot will need to generate a collision-free path to avoid colliding with the pedestrian. This experiment setup is shown in Fig. 14c.
- **Distracted Experiment**: One pedestrian, starting from the left side of the robot, first moves to the right side but quickly shifts back to their initial position. In this case, the robot needs to deal with the situation when the pedestrian suddenly changes their walking direction. The setup is shown in Fig. 14e.

In order to simulate the obstacle avoidance and walking of real pedestrians, we use Reciprocal Collision Avoidance (RVO2) (Li et al. 2019) as the pedestrian simulator in the created simulation environment. We set the preferred speed for robots and pedestrians to be $1.2 m/s$, which is the normal walking speed for people (Montufar et al. 2007).

To evaluate the performance of the proposed method in the dynamic social environment, we compare our approach with three baseline approaches:

- **MPC** (Turri et al. 2013): Autonomous linear MPC controller using simplified models and

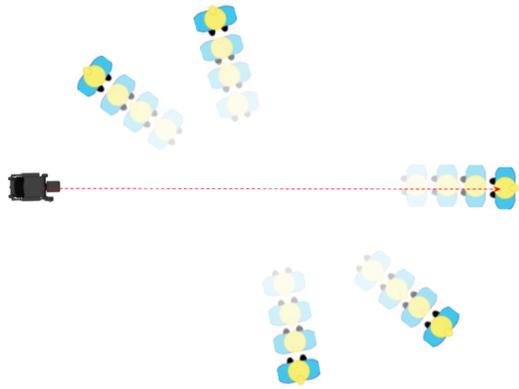

(a) Crossing Experiment Setup

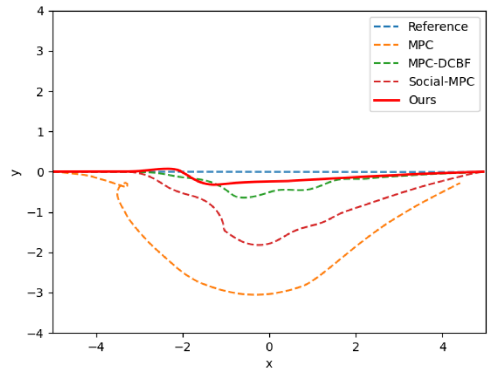

(b) Crossing Experiment Result

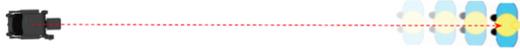

(c) Aggressive Experiment Setup

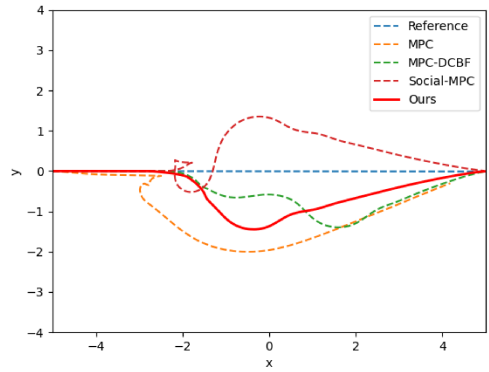

(d) Aggressive Experiment Result

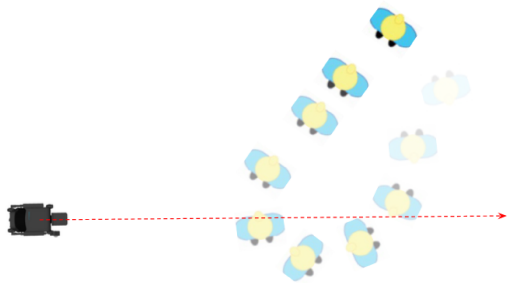

(e) Distracted Experiment Setup

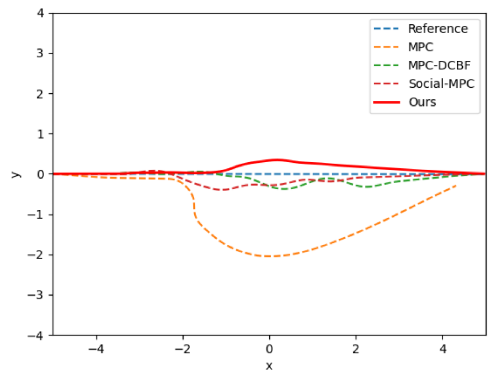

(f) Distracted Experiment Result

**Fig. 14.** The experiment setup and the corresponding qualitative result

considering distance constraints with various Euclidean norms.

- **MPC-DCBF** (Jian et al. 2022): Autonomous MPC tracking controller considering safe constraints with discrete time DCBF.
- **Socially-aware MPC** (Poddar et al. 2023): Socially aware MPC controller considering distance as pedestrians' social area.

To evaluate the safety and efficiency of our Local Planner, we use "Min Dist", "Time", "Traj Length", "Linear Vel Var" and "Angular Vel Var" to compare our algorithm with the three baseline approaches. "Min Dist" represents the minimal distance between the robot and the pedestrians throughout the whole navigation process. "Time" represents the time spent on the navigation process. "Traj Length" represents the trajectory length that the robot takes to the goal. "Linear Vel Var" and "Angular Vel Var" represent the linear and angular velocity variance during the navigation. If the robot generates a winding and less smooth path, the linear and angular velocities tend to have a bigger variance.

For each experiment, the trajectory generated by each algorithm is plotted in Fig. 14b, 14d and 14f. From the plots and the quantitative result shown in TABLE 3a, 3b and 3c, we can conclude that our algorithm generates a most effective and smooth path compared to other three baseline algorithms. The MPC algorithm cannot avoid dynamic obstacles due to the lack of prediction or inaccurate prediction caused by sensor errors. The MPC-DCBF can successfully avoid the dynamic obstacles in all three experiments. However, due to the limitation of the MPC-DCBF in not incorporating the social spaces of both robots and pedestrians, the system, under scenarios simulating distracted or aggressive behavior, generates paths that result in minimal distances to pedestrians, causing potential safety risks.

For socially-aware MPC, although it can generate a path with the maximum distance between the robot and the pedestrians, there are collisions in the aggressive experiment. Another downside for socially-aware MPC is that the path generated in crossing and aggressive experiments is less smooth compared to other experiments, which is reflected in the linear and angular velocity variance. In contrast, our proposed SS-MPC-DCBF Local Planner can smoothly and robustly avoid dynamic

**TABLE 3.** Quantitative results of the dynamic experiments

**(a)** Crossing experiment.

| Method | Min Dis | Time | Traj Length | Linear Vel Var | Angular Vel Var |
|---|---|---|---|---|---|
| MPC | 0m (Collision) | 57.90s | 12.147m | 0.0585 | 0.0852 |
| MPC-DCBF | 0.4898m | **14.72s** | 10.171m | 0.0627 | 0.1799 |
| Socially-aware MPC | **0.8852m** | 15.53s | 11.023m | 0.0557 | 0.2753 |
| SS-MPC-DCBF (Ours) | 0.2411m | 16.89s | **10.067m** | **0.0506** | **0.0741** |

**(b)** Aggressive experiment.

| Method | Min Dis | Time | Traj Length | Linear Vel Var | Angular Vel Var |
|---|---|---|---|---|---|
| MPC | 0m (Collision) | 32.90s | 11.299m | **0.0097** | **0.1770** |
| MPC-DCBF | 0.5102m | **14.84s** | 10.730m | 0.0309 | 0.3591 |
| Socially-aware MPC | 0m (Collision) | 19.76s | 12.737m | 0.0529 | 0.9214 |
| SS-MPC-DCBF (Ours) | **0.5962m** | 15.05s | **10.673m** | 0.0260 | 0.2591 |

**(c)** Distracted experiment.

| Method | Min Dis | Time | Traj Length | Linear Vel Var | Angular Vel Var |
|---|---|---|---|---|---|
| MPC | 0m (Collision) | 29.44s | 10.698m | **0.0079** | 0.0752 |
| MPC-DCBF | 0.2857m | 14.04s | 10.122m | 0.0298 | 0.2369 |
| Socially-aware MPC | 0.2861m | 15.88s | 10.112m | 0.0383 | 0.1844 |
| SS-MPC-DCBF (Ours) | **0.3284m** | 13.755s | **10.018m** | 0.0280 | **0.0513** |

obstacles in advance in these three extreme crowd conditions, exceeding other baseline methods in trajectory length and velocity variance. When the dynamic pedestrians enter the blind point of the LiDAR sensor, the robot can successfully avoid the dynamic obstacles under the user's control.

**Limitations and Future Work**

Our insight is to use shared control to combine the advantages of autonomous navigation and the user's experience, yielding better performance than either alone. However, even though the application of shared control is promising in improving trust and safety, the current input methods and preference-taking from the user, which rely on clicking and tele-operation, can distract the user during navigation in a crowded indoor environment, increasing collision risks. Another issue is that although a real-world experiment was conducted in a tight indoor environment, with the lack of real-world robot data as well as the gap between simulation and real life, the performance of the model is lower than the performance in simulation. Therefore, to replicate the success of the simulation experiments on the real robot, more work on the parameter adjustment of the Local Planner will be needed.

Additionally, our navigation framework builds on the assumption that users can gather sufficient information to decide their preferred route. To enable users to better understand the environment, it is crucial to build a topological semantic map that contains detailed environment information (Hughes et al. 2022) and create a user-friendly Graphical User Interface (GUI) (Baker et al. 2020) for tablets or Virtual Reality (VR) device (Park et al. 2023). These additions will be important considerations when we deploy this framework for experimentation with physical robotic wheelchairs.

**CONCLUSION**

This paper introduced a new shared control-based dynamic navigation system for assistive mobile robots, enhancing user interaction in indoor navigation scenarios. Drawing upon insights from previous studies, our system uniquely incorporates users' path preferences and control inputs into the navigation process. By separating the navigation system into a Global Planner for path selection and a Local Planner for dynamic obstacle avoidance, we address critical limitations in existing autonomous navigation approaches. Our Global Planner innovatively integrates adaptive user path preferences into the path planning algorithm, ensuring the generation of a user-desired collision-free path. This marks a significant step forward in personalizing assistive robotics, enabling users to actively participate in the navigation decision-making process.

Secondly, our system takes a step toward embedding social awareness within the robotic navigation context. By incorporating social areas/spaces into the obstacle modeling process of our Local Planner, our system ensures that respectful distances are maintained from pedestrians, thereby enhancing the robot's ability to navigate socially complex environments. This feature not only improves the robot's operational safety but also its social acceptability, a crucial aspect for widespread adoption in real-world settings.

Thirdly, we propose SS-MPC-DCBF as our Local Planner. This approach enables the generation

of safe, efficient trajectories that avoid potential collisions with moving obstacles in advance, even before they enter a predefined proximity threshold. This predictive capability ensures a higher level of safety and efficiency in the robot's navigation, addressing one of the key challenges in dynamic obstacle avoidance.

Finally, our navigation system addresses the pivotal issue of user distrust in autonomous systems by implementing a shared control mechanism. This approach allows users to dictate their level of involvement in the navigation process, thereby fostering a sense of control and enhancing trust in the system. By empowering users to influence the navigation process, our system not only improves the user experience but also increases the practical applicability of assistive robots in indoor social environments.

In summary, our research advances the field of assistive robotics by proposing a novel navigation framework that prioritizes user preferences, social awareness, predictive obstacle avoidance, and user trust. These advancements pave the way for the development of more intuitive, safe, and user-friendly assistive mobile robots, ultimately enhancing human-robot interaction in navigation and mobility tasks within the built environment. Our project is open-sourced at: https://github.com/Shared-Wheelchair/wheelchair_sim.

**DATA AVAILABILITY STATEMENT**

Some data, models, or code that support the findings of this study are available from the corresponding author upon reasonable request, including the wheelchair simulation model and any Gazebo world file that is used for evaluation.

**ACKNOWLEDGEMENTS**

The work presented in this paper was supported financially by the United States National Science Foundation (NSF) via Award# SCC-IRG 2124857. The support of the NSF is gratefully acknowledged. The authors would also like to thank Jordan Lillie from the Michigan Medicine Wheelchair Seating Service (WSS) for assisting us in assembling all sensors onto the experimental wheelchair platform, and Dennis Gould, also from the WSS, for procuring the experimental wheelchair platform from its manufacturer, Pride Mobility Products for this research.